\documentclass[conference]{IEEEtran}

\usepackage{cite}
\usepackage{amsmath,amssymb,amsfonts}
\usepackage{algorithmic}
\usepackage{graphicx}
\usepackage{textcomp}
\usepackage{xcolor}
\usepackage{multirow}
\usepackage{pifont}
\usepackage[hidelinks]{hyperref}
\def\BibTeX{{\rm B\kern-.05em{\sc i\kern-.025em b}\kern-.08em
    T\kern-.1667em\lower.7ex\hbox{E}\kern-.125emX}}
\begin{document}

\title{Real time dense anomaly detection \\ by learning on synthetic negative data}

\author{\IEEEauthorblockN{Anja Delić, Matej Grcić, Siniša Šegvić}
\textit{Master research seminar at University of Zagreb Faculty of Electrical Engineering and Computing}}

\maketitle

\begin{abstract}
Most approaches to dense anomaly detection rely on generative modeling or on discriminative methods that train with negative data. 
We consider a recent hybrid method that optimizes the same shared representation according to cross-entropy of the discriminative predictions, and negative log likelihood of the predicted energy-based density. 
We extend that work with a jointly trained generative flow that samples synthetic negatives at the border of the inlier distribution.
The proposed extension provides potential to learn the hybrid method without real negative data.
Our experiments analyse the impact of training with synthetic negative data and validate contribution of the energy-based density during training and evaluation.
\end{abstract}


\section{Introduction}
Existing deep learning methods solve the computer vision task of semantic segmentation with high precision. However, models are typically evaluated on closed-world datasets with a fixed taxonomy and their behaviour in real-world applications can be unpredictable. 
We therefore study methods that detect objects that deviate from the training distribution by assigning a pixel vise anomaly score. This task has also been known as dense anomaly, out-of-distribution (OoD) or novelty detection. 
Current state of the art in this field relies on training on mixed content images obtained by pasting negative data atop inlier images. 
However, anomaly detection methods that utilize real negative datasets can give rise to over-optimistic performance metrics because of the bias introduced by the possibility that some pasted negatives can be related to anomalies in the test dataset.
We try to avoid this bias by extending discriminative \cite{bevandic19gcpr} \cite{bevandic22arxiv} and hybrid \cite{grcic22eccv} methods that utilize real negatives with the capability to learn with generated synthetic negative data \cite{grcic21arxiv, grcic21visapp}. 
This work analyzes methods that were proposed in \cite{grcic23arxiv}. We extend that work by providing real-time experiments on the StreetHazards dataset \cite{hendrycks19icml}.

\section{Related work} 
\label{sec:realted-work}
Anomaly detection methods can be discriminative or generative. 
Most discriminative methods require training with auxiliary negative data that is pasted over regular training images\cite{bevandic19gcpr, bevandic22arxiv}.
Pixel-wise OoD predictions can be produced by a separate OoD head sharing features with the standard semantic segmentation head or by measuring uncertainty of the predictive distributions.
Real negative data can be replaced with synthetic negative patches obtained by sampling a jointly trained generative model\cite{lee18iclr, neal18eccv, zhao21arxiv, grcic21arxiv, grcic21visapp}.

Generative methods detect outliers according to the estimated density of the generative distribution of the inlier training data. The data likelihood has been estimated by leveraging  an energy-based model\cite{grcic22eccv} or a normalizing flow\cite{blum2021fishyscapes, dinh17iclr}.
However, estimating the data likelihood with a generative model can be suboptimal because the generative model may assign higher likelihood to outliers than inliers\cite{nalisnick19iclr, kirichenko20neurips}.
On the other hand, discriminative methods highly depend on the choice of negative data. 
A recent hybrid method known as  DenseHybrid \cite{grcic22eccv} aims at  achieving a synergy between these two approaches. It builds on a recent image-wide approach \cite{grathwohl20iclr}.
DenseHybrid\cite{grcic22eccv} combines discriminative and generative learning by reinterpreting discriminative logits as a logarithm of unnormalized joint distribution of data input and labels. 




\section{Learning with synthetic negatives}
\label{sec:anomaly-detection}
We extend the training procedure proposed od DenseHybrid to jointly train a normalizing flow (NF) that generates samples at the border of the inlier manifold \cite{lee18iclr, grcic21visapp}. 

For the segmentation model we use the original DenseHybrid loss \cite{grcic22eccv} that combines standard segmentation loss $L_{cls}$, outlier head loss $L_{d}$ and energy loss $L_{x}$:
\begin{equation}
\label{eq:dh-loss}
\begin{aligned}
    L_{\mathrm{seg}}(\theta,\gamma)=&-\mathbb{E}_{\mathbf{x},\mathbf{y}\in D_{\text{in}}}[\ln P(\mathbf{y} \mid \mathbf{x})+\ln P(d_{\text{in}} \mid \mathbf{x})] \\ 
    & -\beta \cdot \mathbb{E}_{\mathbf{x} \in D_{\text{out}}}[\ln (P(d_{\text{out}} \mid \mathbf{x}))-\ln \hat{p}(\mathbf{x})].
\end{aligned}
\end{equation}

We try two loss functions for training the NF. The first approach corresponds to the Jensen-Shannon divergence between $P(y|\mathbf{x})$ and the uniform distribution to the negative log-likelihood of the cut out inlier patches: 
\begin{equation}
    \label{eq:flow-loss}
    L_{\mathrm{flow}}(\psi)= L_{\mathrm{mle}}(\psi) + \beta \cdot L_{\mathrm{jsd}}(\psi, \theta).
\end{equation}
The first term attracts the generative distribution towards inliers while the second term pushes it away from the inliers. Thus, the normalizing flow generates data at the boundary of the training distribution. 
We stop the gradients from $L_{d}$ and $L_{x}$ from backpropagating to the NF so that only gradients from \eqref{eq:flow-loss} impact the NF. We refer to this method as NF-Hybrid-JS.

In the second approach, we remove the JS divergence from the NF training objective \eqref{eq:flow-loss} and let the gradients from $L_{d}$ and $L_{x}$ to propagate through the segmentation model back to the NF. We refer to this method as NF-Hybrid-LdLx.
\begin{equation}
    \label{eq:mle}
    L_{\mathrm{mle}}(\psi)=-\mathbb{E}_{\mathbf{\operatorname{crop}(x)}\in D_{\text{in}}}[\ln p_\psi(\operatorname{crop}(\mathbf{x}))]
\end{equation}

\begin{table*}[htb!]
    \caption{Anomaly detection, closed-set and open-set performance on StreetHazards \cite{hendrycks19icml}}
    \begin{center}
    \begin{tabular}{*{8}{c}}
         Method & Aux. data & Anomaly score & AP & $\text{FPR}_{95}$ & AUROC & closed $\overline{\text{IoU}}$ & open $\overline{\text{IoU}}$ \\
         \hline\hline
         \noalign{\smallskip}
         \multirow{2}{*}{DenseHybrid \cite{grcic22eccv}} & \multirow{2}{*}{\ding{51}} & DH & 23.81 & 18.76 & 93.69 & 57.52 & 37.59\\ & & OP$\times$MS & 17.08 & 21.94 & 92.13 &  57.52 & 37.39\\
         \hline\noalign{\smallskip}
         NFlowJS \cite{grcic21arxiv} & \ding{55} & JSD & 12.12 & 25.64 & 90.81 & 60.34 & 42.94 \\
         \hline\noalign{\smallskip}
         NF-Hybrid-JS & \ding{55} & DH & 16.61 & 21.76 & 91.83 & 56.74 & 35.72 \\
         \noalign{\smallskip}
         \multirow{2}{*}{NF-Hybrid-LdLx} & \multirow{2}{*}{\ding{55}} 
         & DH & 16.67 & 19.26 & 92.39 & 56.76 & 37.88 \\
         && OP$\times$MS & 14.35 & 23.03 & 90.96 & 56.76 & 34.57 \\
         \noalign{\smallskip}
         NF-Hybrid-Ld & \ding{55} & DH & 15.62 & 22.78 & 91.31 & 56.79 & 35.07 \\
          \hline\noalign{\smallskip}
         \multirow{3}{*}{OODHead} & \multirow{2}{*}{\ding{51}} 
         & DH & 24.63 & 16.45 & 94.51 & 58.51 & 44.29 \\
         && OP & 22.46 & 31.26 & 91.71 & 58.51 & 32.78 \\
         && OP$\times$MS & 22.52 & 19.06 & 93.54 & 58.51 & 40.13 \\
         \hline\noalign{\smallskip}
         \multirow{3}{*}{NF-OODHead} & \multirow{3}{*}{\ding{55}} 
         & DH & 12.13 & 21.73 & 92.03 & 60.11 & 41.76 \\
         && OP & 3.83 & 69.12 & 70.03 & 60.11 & 9.8 \\
         && OP$\times$MS & 8.87 & 23.99 & 90.55 & 60.11 & 36.92 \\
         \hline
    \end{tabular}
    \label{tab:results}
\end{center}
\end{table*}

We validate the influence of the energy loss in \eqref{eq:dh-loss} by removing it from the segmentation model training objective. That results in a discriminative segmentation model with an additional outlier head similar to approaches presented in \cite{bevandic19gcpr} \cite{bevandic22arxiv}. We present experiments with real (method OODHead) and synthetic negatives (NF-OODHead). When using synthetic negatives, the normalizing flow is trained by minimizing negative log-likelihood \eqref{eq:mle}.

\section{Anomaly Score}
\label{sec:anomaly-score}
During inference the anomaly detector recognizes anomalies by thresholding a predefined anomaly score. A simple approach is to define the anomaly score as a discriminative probability that the sample is an outlier\cite{bevandic19gcpr}:
\begin{equation}
    \label{eq:score-d}
    s_{OP}(\mathbf{x}) = P(d_{\text{out}}=1 \mid \mathbf{x}).
\end{equation}
Performance on small anomalous objects can be improved by multiplying the outlier probability with max-softmax\cite{bevandic22arxiv}:
\begin{equation}
    \label{eq:score-msp}
    s_{OP\times MS}(\mathbf{x}) = P(d_{\text{out}}=1 \mid \mathbf{x}) \cdot (1-\operatorname{max}_{c}(P(y_{c}|\mathbf{x}))).
\end{equation}
The DenseHybrid anomaly score is defined as the ratio of dataset posterior and data likelihood\cite{grcic22eccv}: 
\begin{equation}
    \label{eq:score-dh}
    s_{DH}(\mathbf{x}) = \ln\frac{P(d_{\text{out}} \mid \mathbf{x})}{p(\mathbf{x})}.
\end{equation}
Anomaly score can also be defined as JS divergence between $P(y|\mathbf{x})$ and the uniform distribution\cite{grcic21arxiv}. In later text we refer to these methods as outlier probability (OP), outlier probability multiplied by maximum softmax (OP$\times$MS), DenseHybrid score (DH) and JS divergence score (JSD).

\section{Experiments}
\label{sec:experiments}

We evaluate methods described in Section \ref{sec:anomaly-detection} on StreetHazards dataset \cite{hendrycks19icml}. 
In all our experiments, we use SwiftNet-RN18 \cite{orsic2019defense} as the segmentation model because it shows good semantic segmentation performance with real-time inference.
We create mixed content images by pasting negatives atop inlier crops obtained by randomly jittering in range [0.5, 2], random horizontal flipping and taking a square crop of size 768. 
In experiments that utilize real negative data, we paste instances from ADE20k dataset \cite{zhou2017scene}. We use DenseFlow-25-6\cite{grcic21neurips} for generating synthetic negatives. 
The normalizing flow generates rectangular patches with arbitrary spatial dimensions. In experiments we use square dimensions between 16 and 216 pixels. The NF pretrained on Vistas is trained with the Adamax optimizer with learning rate $10^{-6}$.
The joint training is a two step process. For initialization we use weights pretrained on ImageNet. We first train our model on StreetHazards train for 80 epochs as a standard segmentation model. Then we fine tune for additional 40 epochs on mixed content images following the anomaly aware procedures as described in Section \ref{sec:anomaly-detection}. We use Adam optimizer with initial learning rate $10^{-5}$ that is decayed through a cosine schedule to $10^{-7}$.
We set the loss hyperparameter $\beta$ for $L_{x}$ and $L_{JSD}$ to 0.03 and for $L_{d}$ 0.3. In OODHead and NF-OODHead experiments we use $\beta = 1$. We train with batch size 16.
We evaluate our experiments with different anomaly scores as described in Section \ref{sec:anomaly-score}. We use temperature scaling with temperature $T=2$ for OP, OP$\times$MS and JSD scores. 

Table \ref{tab:results} presents the experimental performance. To make our results comparable with previous work, we reproduce experiments \cite{grcic22eccv} and \cite{grcic21arxiv}, DenseHybrid and NFlowJS respectively, following the original setup but with SwiftNet-RN18 as the semantic segmentation model. We evaluate DenseHybrid with the proposed DH score and with OP$\times$MS to be comparable with our experiments. We see a performance drop when using OP$\times$MS.

The NF-Hybrid-JS approach combines DenseHybrid and NFlowJS. The difference to NFlowJS is an improvement in training the segmentation model by adding the energy term and an outlier head in DenseHybrid manner. We use the DH anomaly score in evaluation. NF-Hybrid-JS outperforms NFlowJS but not DenseHybrid. 

The NF-Hybrid-LdLx approach achieves the best average precision score in comparison to our other experiments that use synthetic negatives and NFlowJS. We stop the gradients from the energy loss from propagating to the NF (method NF-Hybrid-Ld) so that only the outlier head influences the NF but this leads to worse performance. This suggests that the energy loss can be beneficial for training the NF.

\begin{table}[htb!]
    \caption{Computational overhead in frames per second of used anomaly detectors when inferring with RTX A4000 on StreetHazards test images}
    \begin{center}
    \begin{tabular}{*{2}{c}}
         Method & FPS \\
         \hline
         \noalign{\smallskip}
         OH & 74.43\\
         OP$\times$MS  &  71.69\\
         DH & 69.86 \\
         JSD & 52.41 \\
    \end{tabular}
    \label{tab:fps}
\end{center}
\end{table}

OODHead experiment is an ablation of the DenseHybrid approach. A standard approach for evaluating a two head model would be to use OP anomaly score. We also evaluate OP$\times$MS and DH. OP$\times$MS outperforms OP and is best seen in a big drop in FPR metric. DH score outperforms both OP$\times$MS and OP. 
It is interesting that OODHead outperforms DenseHybrid which could be caused by insufficient capacity of the segmentation model. 
When using DH score we see that we can improve the discriminative anomaly aware model by using the data likelihood only in evaluation.

We replace the auxiliary negatives with synthetic negatives generated by a normalizing flow (NF-OODHead) and evaluate using OP, OP$\times$MS and DH anomaly scores. We see that this method highly depends on the choice of anomaly score. There is a noticeable performance gap between OP and OP$\times$MS scores. 

We measure open-set performance using the open-IoU metric proposed in \cite{grcic22eccv}. If we compare experiments that differ only in using the energy term in the loss function, DenseHybrid and OODHead or also NF-Hybrid-JS and NFlowJS, we can observe a drop in both open and closed IoU when learning in a hybrid manner. This indicates that the segmentation model SwiftNetRN-18 does not have enough capacity for hybrid learning. 

Table \ref{tab:fps} shows computational overhead for the open set recognition task over the SwiftNetRN-18 segmentation model for anomaly detectors described in \ref{sec:anomaly-score}. We measured the inference time with NVIDIA RTX A4000 on StreetHazards test images of size 720 $\times$ 1280 pixels.

\section{Conclusion}
We extend a hybrid dense anomaly detection approach with a normalizing flow model that generates synthetic negatives following methods that were proposed in \cite{grcic23arxiv}. 
This allows to learn outlier-aware models without training on real negative data. Our experiments show improvement of hybrid approach over other methods that use synthetic negatives. However, we observed a performance drop when training with the hybrid approach on real negative data which indicates that the chosen semantic segmentation model SwiftNetRN-18 does not have sufficient capacity for hybrid learning. 
We show that the normalizing flow can generate meaningful anomalies when influenced by the energy loss and outlier head. We evaluate our training approaches using different anomaly scores and show advantage of the hybrid anomaly score even when the segmentation model is trained in a discriminative manner. These findings should be validated on larger models and more datasets. Interesting directions for future work include reducing the gap between closed and open set performance and considering different anomaly detection approaches.

\bibliographystyle{IEEEtran}

\begin{thebibliography}{00}
\bibitem{bevandic19gcpr} P. Bevandić, I. Krešo, M. Oršić, and S. Šegvić, ``Simultaneous semantic
segmentation and outlier detection in presence of domain shift,'' in 41st
DAGM German Conference, DAGM GCPR. Springer, 2019.

\bibitem{bevandic22arxiv} P. Bevandić, I. Krešo, M. Oršić, and S. Šegvić, ``Dense open-set recognition based on training with noisy negative images,'' Image and Vision Computing, 2022.

\bibitem{grcic22eccv} M. Grcić, P. Bevandić, and S. Šegvić, ``DenseHybrid: Hybrid Anomaly Detection for Dense Open-Set Recognition,'' in European Conference on Computer Vision, ECCV 2022. Springer, 2022.

\bibitem{grcic21arxiv} M. Grcić, P. Bevandić, and S. Šegvić, ``Dense anomaly detection by robust learning on synthetic negative data,'' arxiv preprint
arXiv:2112.12833, 2021.

\bibitem{grcic21visapp} M. Grcić, P. Bevandić, and S. Šegvić, ``Dense open-set recognition with synthetic outliers generated by real NVP,'' in 16th International Joint Conference on Computer Vision,Imaging and Computer Graphics Theory and Applications, VISIGRAPP, 2021.

\bibitem{grcic23arxiv} M. Grcić, P. Bevandić, and S. Šegvić, ``Hybrid Open-set Segmentation with Synthetic Negative Data,'' arXiv preprint
arXiv:2301.08555, 2023.

\bibitem{hendrycks19icml} D. Hendrycks, S. Basart, M. Mazeika, A. Zou, J. Kwon, M. Mostajabi,
J. Steinhardt, and D. Song, ``Scaling out-of-distribution detection for real-world settings,'' ICML, 2022.

\bibitem{lee18iclr} K. Lee, H. Lee, K. Lee, and J. Shin, ``Training Confidence-calibrated Classifiers for Detecting Out-of-Distribution Samples,'' in International
Conference on Learning Representations, 2018.

\bibitem{neal18eccv} L. Neal, M. Olson, X. Fern, W.-K. Wong, and F. Li, ``Open set learning
with counterfactual images,'' in Proceedings of the European Conference on Computer Vision (ECCV), 2018.

\bibitem{zhao21arxiv} Z. Zhao, L. Cao, and K. Lin, ``Revealing distributional vulnerability of
explicit discriminators by implicit generators,'' CoRR, 2021.

\bibitem{blum2021fishyscapes} H. Blum, P.-E. Sarlin, J. Nieto, R. Siegwart, and C. Cadena, ``The fishyscapes benchmark: Measuring blind spots in semantic segmentation,'' International Journal of Computer Vision, 2021.

\bibitem{dinh17iclr} L. Dinh, J. Sohl-Dickstein, and S. Bengio, ``Density estimation using
real NVP,'' in International Conference on Learning Representations,
2017.

\bibitem{nalisnick19iclr} Nalisnick, A. Matsukawa, Y. W. Teh, D. Gorur, and B. Lakshminarayanan, ``Do deep generative models know what they don’t know?,'' in International Conference on Learning Representations, 2019.

\bibitem{kirichenko20neurips} P. Kirichenko, P. Izmailov, and A. G. Wilson, ``Why normalizing flows
fail to detect out-of-distribution data,'' in Advances in Neural Information
Processing Systems, 2020.

\bibitem{grathwohl20iclr} W. Grathwohl, K.-C. Wang, J.-H. Jacobsen, D. Duvenaud, M. Norouzi,
and K. Swersky, ``Your classifier is secretly an energy based model and
you should treat it like one,'' in International Conference on Learning
Representations, 2020.

\bibitem{orsic2019defense} M. Orsic, I. Kreso, P. Bevandic, and S. Segvic, ``In defense of pre-trained imagenet architectures for real-time semantic segmentation of road-driving images,''  in Proceedings of the IEEE/CVF Conference on
Computer Vision and Pattern Recognition, 2019, pp. 12 607–12 616.

\bibitem{zhou2017scene} B. Zhou, H. Zhao, X. Puig, S. Fidler, A. Barriuso, and A. Torralba, ``Scene parsing through ade20k dataset,'' in Proceedings of the IEEE
conference on computer vision and pattern recognition, 2017.

\bibitem{grcic21neurips} M. Grcić, I. Grubišić, and S. Šegvić, ``Densely connected normalizing flows,'' Advances in Neural Information Processing Systems, 2021.

\end{thebibliography}

\vspace{12pt}

\end{document}